\documentclass[letterpaper, 10 pt, conference]{ieeeconf}

\IEEEoverridecommandlockouts
\usepackage{booktabs}
\usepackage{tabularx}
\usepackage[table]{xcolor}
\colorlet{oursrow}{blue!8}
\usepackage{cite}
\usepackage{amsmath,amssymb,amsfonts}
\usepackage{graphicx}
\usepackage{multirow}
\usepackage{pifont}
\usepackage{xcolor}
\usepackage{cuted}
\usepackage{capt-of}

\usepackage{url}
\makeatletter
\let\NAT@parse\undefined
\makeatother
\definecolor{linkblue}{HTML}{1F4E79}
\usepackage[colorlinks=true,allcolors=linkblue,urlcolor=magenta]{hyperref}

\newcommand{\qvec}{\mathbf{q}}

\newcommand{\humanoidPose}{HumanoidPose}

\newcommand{\myparagraph}[1]{\noindent\textbf{#1}}

\title{\LARGE \bf
Vision as Proprioception: Estimating Humanoid Robot Pose from Video}

\author{Jiaxi Jiang\hspace{1.5em}Matthias Heyrman\hspace{1.5em}Ren\'e Zurbr\"ugg\hspace{1.5em}Marco Hutter\hspace{1.5em}Christian Holz%
\\
{\normalsize\textbf{ETH Z\"urich}}%
\\
{\normalsize\url{https://github.com/eth-siplab/HumanoidPose}}%
}

\makeatletter
\def\@IEEEdynamictitlevspace{\vspace{-8pt}}
\makeatother

\makeatletter
\long\def\@makecaption#1#2{%
\ifx\@captype\@IEEEtablestring%
\begin{center}{\footnotesize #1}\\{\footnotesize #2}\end{center}%
\@IEEEtablecaptionsepspace%
\else
\@IEEEfigurecaptionsepspace%
\setbox\@tempboxa\hbox{\footnotesize #1.~~ #2}%
\ifdim \wd\@tempboxa >\hsize%
\setbox\@tempboxa\hbox{\footnotesize #1.~~ }%
\parbox[t]{\hsize}{\footnotesize \noindent\unhbox\@tempboxa#2}%
\else%
\ifcenterfigcaptions \hbox to\hsize{\footnotesize\hfil\box\@tempboxa\hfil}%
\else \hbox to\hsize{\footnotesize\box\@tempboxa\hfil}%
\fi\fi\fi}
\makeatother

\usepackage{caption}
\begin{document}
\maketitle
\thispagestyle{empty}
\pagestyle{empty}

\begin{strip}
\centering
\includegraphics[width=\textwidth]{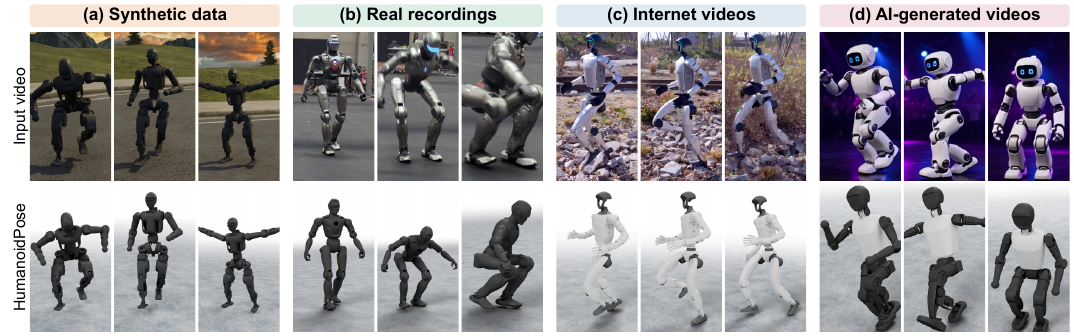}
\captionof{figure}{We present \textbf{\humanoidPose{}}, a new framework that estimates full-body 3D humanoid robot configurations from video alone.
Input frames (top) and recovered robot configurations (bottom) are shown for (a)~synthetic data, (b)~real-robot recordings, (c)~internet videos, and (d)~AI-generated videos.
HumanoidPose is a single shared model that supports multiple robot embodiments without internal state measurements, providing a basis for robot interaction and learning from video.
}
\label{fig:teaser}
\end{strip}

\begin{abstract}
Estimating a humanoid robot’s internal state from video has a wide range of real-world applications but remains unexplored. During interactions with a robot, its onboard state information may be unavailable due to privacy, ownership, or communication constraints.
Meanwhile, online and AI-generated robot videos provide rich sources of data but typically lack corresponding joint configurations, limiting their direct use in robot learning.
To address these issues, we present \humanoidPose{}, a new framework that treats vision as proprioception to recover full-body joint configurations in world space from video alone.
Specifically, we combine pretrained human-pose features with robot kinematics in a unified model that predicts each embodiment's joint configurations.
We retarget human motions to sixteen different humanoid embodiments, enabling scalable learning entirely from synthetic data.
To improve real-world generalization, we further leverage unlabeled online videos through self-training to develop a 2D robot-pose detector.
In our evaluation, \humanoidPose{} improves joint-angle and joint-position accuracy over the evaluated baselines and reduces joint-angle error by $18$--$32\%$ relative to the best-performing baseline on each of the three real humanoid platforms.
We further show improved global localization and generalization to unseen embodiments.
\humanoidPose{} enables visual understanding of other robots during interaction and extraction of robot motion from online and generated videos.
\end{abstract}

\section{Introduction}
Understanding and interacting with humanoid robots requires interpreting how their bodies move. People and other robots may need to anticipate a reaching arm, recognize an action, or coordinate with the robot. Although onboard sensors measure the robot’s configuration, privacy, ownership, or communication constraints may prevent access to these measurements, making vision the primary means for an external observer to recover it. The same need extends to learning from video: online robot videos capture motion in varied environments without accompanying joint-state logs, while AI-generated videos provide candidate motion demonstrations with no physical robot or encoder measurements at all. Together with real-robot recordings and synthetic videos, these sources motivate recovering articulated robot motion directly from visual observations (Fig.~\ref{fig:teaser}). Expressing this motion in the robot’s joint coordinates makes it available for analysis, prediction, and learning.

We study \emph{visual proprioception for humanoid robots:} recovering full-body joint configurations in world space from ordinary video without access to onboard state. This setting introduces several challenges. Humanoids differ in body proportions, joint layouts, and mechanical limits, requiring an estimator that can accommodate different kinematic structures and output spaces. Occlusion and monocular depth ambiguity make articulation underdetermined from appearance alone, calling for learned priors that generalize across embodiments, motions, and visual conditions. Recovering world-space trajectories further requires separating robot motion from camera motion.

Learning these priors requires large-scale data pairing images with robot configurations
across motions, embodiments, and environments. Such data are difficult to obtain:
collecting encoder-synchronized footage across many platforms is costly.
Existing robot pose estimators address
articulated manipulators~\cite{robopose,horopose,robopepp}, but humanoid motion recovery
requires full-body estimation across robot designs and appearances.
Human mesh recovery offers a way to reuse priors learned from extensive human
imagery~\cite{hmr2,promlthmr}: estimate a human pose and retarget it to the
robot~\cite{gmr}. However, human articulation differs from robot articulation,
and retargeting can amplify prediction errors. We instead transfer pretrained human-pose
features to directly predict robot joint poses.

We present \textbf{HumanoidPose}, a framework for recovering full-body robot pose
and world-space motion from video. Its key design is a shared decoder that combines
frozen human-pretrained image features with queries derived from robot kinematics to
predict joint configurations and root pose across embodiments. We combine the
predicted root pose with estimated camera motion to recover world-space trajectories.
Queries follow each robot's joint layout, allowing a shared decoder and prediction head.

Our synthetic pipeline retargets motions from the Archive of Motion Capture as
Surface Shapes (AMASS)~\cite{amass} to sixteen humanoids using General Motion
Retargeting (GMR)~\cite{gmr} and renders paired images and robot states under varied
conditions. These data supervise the pose decoder and initialize a YOLO11-based detector
for robot boxes and 2D keypoints, and unlabeled online videos further adapt this detector. Across three real platforms,
the unified model improves joint-angle and joint-position accuracy over the
evaluated baselines.

\noindent Our contributions are:
\begin{itemize}
    \item We formulate visual proprioception for humanoid robots, the problem of recovering full-body joint configurations and world-space motion from video without access to onboard state.
    \item We introduce a unified, kinematically conditioned model for 3D robot pose estimation that predicts joint configurations and root pose across humanoid embodiments without robot-specific prediction heads.
    \item We generate $128{,}000$ synthetic images across sixteen humanoids with paired 2D and 3D labels to train the 3D pose model and a 2D detector, further adapting the detector through self-training on online videos.
    \item We evaluate HumanoidPose on three real humanoid platforms, demonstrating improved joint-angle and joint-position accuracy, improved root translation on G1, and transfer and adaptation to unseen embodiments.
\end{itemize}

\begin{figure*}[t]
\vspace{5pt}
\centering
\includegraphics[width=\textwidth]{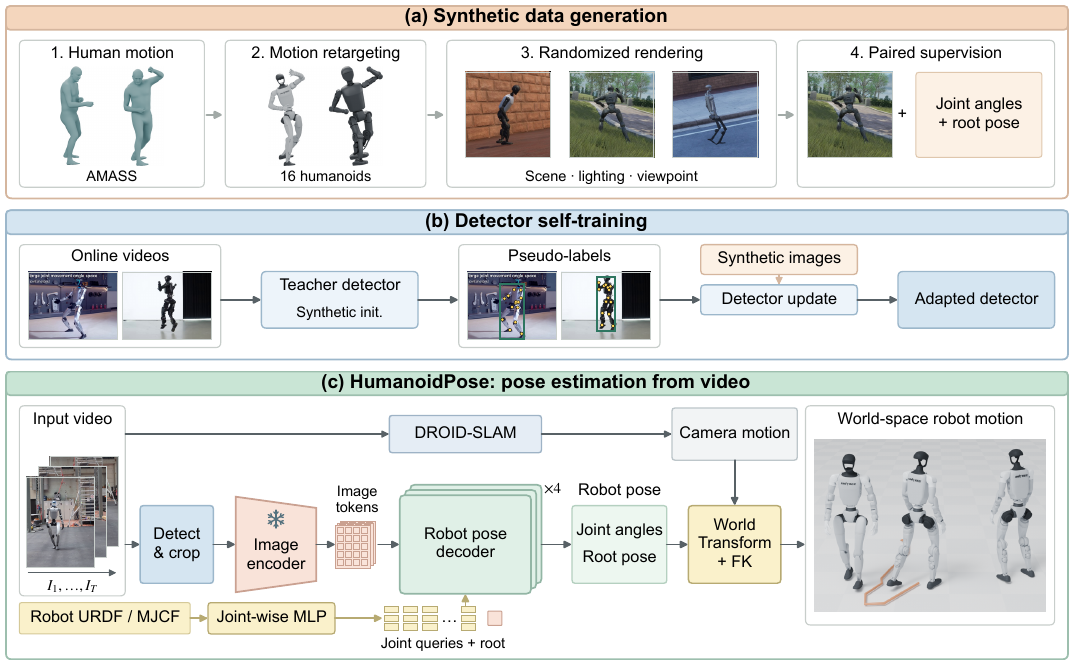}
\caption{HumanoidPose overview. (a) Retargeted human motions and randomized
rendering provide synthetic robot pose supervision. (b) Pseudo-labels from unlabeled
online videos adapt the detector through self-training. (c) Frozen human-pretrained
features and kinematic queries drive a shared decoder for joint angles and root pose.
Camera motion and forward kinematics place predictions in world coordinates.
The pose decoder is trained solely on synthetic data.}
\label{fig:pipeline}
\vspace{-5mm}
\end{figure*}
\section{Related Work}

\noindent\textbf{Robot Pose Estimation.} Camera-to-robot calibration methods such as
DREAM and PoseFusion combine image keypoints with known 3D robot keypoint locations to
estimate the camera transform~\cite{dream,ctrnet,posefusion}. Recovering unknown joint
states additionally requires estimating articulation, addressed through rendering-based
refinement in RoboPose and GISR~\cite{robopose,gisr,diffrobotrender}, geometric lifting~\cite{distancegeom,robokeygen},
and feed-forward regression in HoRoPose~\cite{horopose}. More recent methods such as
RoboPEPP and RoboTAG learn visual representations that reflect robot articulation~\cite{robopepp,ropes,robotag}.
For humanoids, Amini et al. and Heo et
al.~\cite{amini2021,heo2024} estimate 2D image keypoints across robot platforms, without
recovering full 3D articulation or actuated joint configurations. In contrast, HumanoidPose
estimates full-body 3D humanoid robot poses in robot coordinates across embodiments.

\noindent\textbf{Human Pose and Motion Estimation.} Parametric body models and image
fitting~\cite{smpl,smplify,smplx} underpin learned body recovery~\cite{hmr,spin,pare}, while
kinematic formulations connect joint positions and body rotations~\cite{dkpr,hybrik,manikin}.
SAM~3D~Body~\cite{sam3dbody} supports keypoint and mask prompts for single-image body
recovery. Video methods progress from temporal motion priors~\cite{vibe} to global motion
reconstruction~\cite{glamr,slahmr}. WHAM and TRAM~\cite{wham,tram} explicitly account for
camera motion, while GVHMR~\cite{gvhmr} uses gravity-view coordinates.
PromptHMR~\cite{promlthmr} incorporates spatial and semantic prompts into full-image
SMPL-X recovery and extends to temporally coherent reconstruction from video.
Complementary approaches combine egocentric tracking and exocentric observations
for full-body motion capture~\cite{egoexomocap}.
However, humanoid morphology, joint limits and actuated coordinates make human-body
representations indirect for robot-state recovery. \humanoidPose{} directly estimates
robot joint coordinates and world-space motion from video.

\noindent\textbf{Synthetic Data and Sim-to-Real Transfer.} Synthetic human data~\cite{surreal} and
domain randomization~\cite{domainrand} establish rendering as a source of visual supervision.
Motion archives~\cite{amass} and increasingly realistic datasets~\cite{agora,bedlam,synbody,egosim,bedlam2}
broaden coverage of appearance, motion, and scenes, while procedural
tools~\cite{blenderproc} automate rendering and annotation. Rendered robot images and adaptation
techniques also support transfer to real robot observations~\cite{dream,robopose,ctrnet,heo2024}.
Motion retargeting~\cite{nkn} further connects human motion resources to new embodiments,
supporting humanoid control and imitation~\cite{exbody,h2o,humanplus,omnih2o}. Recent pipelines
such as UH-1 and GMR retarget human motion capture and internet videos into humanoid robot references~\cite{uh1,gmr},
showing that human motion priors can provide scalable supervision for humanoid behaviors.
We combine retargeted motions with robot-specific rendering to pair images with the exact joint
configurations used to generate them. This directly supervises actuated coordinates, enabling
our decoder to train entirely on synthetic images and transfer to real robot videos.

\section{Method}

\subsection{Problem Formulation and Overview}
\humanoidPose{} takes an external red--green--blue (RGB) video $\{I_t\}_{t=1}^{T}$
and a kinematic description in Unified Robot Description Format (URDF) or MuJoCo's
native model format (MJCF). For a robot with $n$ actuated joints, each frame yields
a configuration $\hat{\qvec}_t\in\mathbb{R}^{n}$ and a root-to-camera transform
$\hat{\mathbf{T}}_{C\leftarrow R,t}$. Root estimation uses camera intrinsics and
nominal metric robot height. Combining predictions with camera motion recovers
world-space trajectories.

Figure~\ref{fig:pipeline} shows synthetic supervision, detector adaptation and pose decoding.
At inference, the adapted YOLO11 detector crops the robot. A frozen human-pretrained
encoder extracts crop features, which the shared decoder combines with kinematic
joint queries to predict joint configurations and root pose.
Temporal information supports detector self-training and camera estimation.
Pose decoding is per-frame.

\subsection{Training Data and Detector Adaptation}
\label{sec:data}
\myparagraph{Synthetic data generation.}
Our synthetic pipeline turns human motion archives into paired robot images and states
across diverse mechanisms. We first retarget AMASS human motions~\cite{amass} to sixteen
humanoid URDF/MJCF models using GMR~\cite{gmr}, producing robot joint trajectories and
root poses in each embodiment's own coordinates. We render these configurations
in Unreal Engine~\cite{unrealengine} with randomized scenes, lighting, camera viewpoints and robot
appearances~\cite{domainrand}. For each rendered frame, we project a shared set of
robot body landmarks through the rendering camera to obtain 2D keypoint annotations
and derive bounding boxes for detector training. Each RGB image is paired with its
camera intrinsics, root pose, exact joint configuration and corresponding 2D annotations.
These pairs supervise robot configurations and 2D pose across different heights,
proportions and joint counts. A common compiled robot model defines coordinate
conventions, joint limits and forward kinematics (FK), keeping labels, descriptors and
training losses consistent.

\begin{figure*}[t]
\vspace{5pt}
\centering
\includegraphics[width=\linewidth]{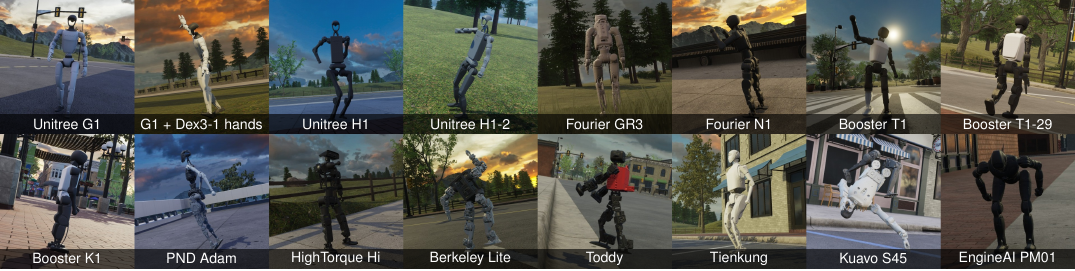}
\caption{The training corpus. Human motion capture retargeted onto sixteen humanoids
spanning $3.0\times$ in height ($0.60$--$1.78$\,m) and $19$--$43$ degrees of freedom, rendered
in Unreal Engine under randomized scene, illumination and viewpoint. One sample is shown
per robot, paired with the configuration used to render it.}
\label{fig:data}
\vspace{-10pt}
\end{figure*}

\myparagraph{2D robot pose estimation and self-training.}
\label{sec:detector}
To localize humanoids across embodiments, we finetune a YOLO11 pose
model~\cite{yolo11}, initialized from human keypoint weights, on the paired images
and 2D annotations generated by our synthetic pipeline. The model learns to predict
a robot bounding box and $15$ shared body landmarks: pelvis, neck, head, and the
left and right hips, knees, feet, shoulders, elbows and wrists. This common landmark
definition supports robots with different actuated joint layouts.

We further adapt the model to real-world appearances through self-training on
unlabeled online videos. Confident detections with sufficient visible keypoints
serve as temporal anchors, and expanded keypoint extents define their bounding
boxes. We fill short detection gaps by interpolating jointly visible keypoints
between nearby anchors, retaining only single-robot pairs with overlapping boxes
and excluding ambiguous detections. The resulting pseudo-labeled frames are mixed
with synthetic images to finetune the model. We repeat pseudo-label generation and
training for two rounds, excluding evaluation recordings. The 3D decoder trains
solely on synthetic poses.

At inference, the adapted model localizes the robot and defines a square RGB crop
with robot-height fraction $\rho=0.55$. The 3D decoder predicts robot pose from the
crop's visual features and robot descriptors. The predicted 2D landmarks also
provide the observations for the geometric fitting baseline in our experiments.

\subsection{Kinematics-Conditioned 3D Pose Estimation}
\label{sec:tokens}
\label{sec:limits}
\myparagraph{Joint queries.} A shared encoder maps each actuated joint's description
to a query, allowing one decoder to handle variable joint layouts. Building on per-joint
representations in morphology-conditioned control~\cite{metamorph,urma,getzero},
each query encodes geometry, motion range and kinematic-tree position.

For joint $j$, let $\mathbf{a}_j$ and $\mathbf{c}_j$ denote its axis and anchor in
the local body frame, and $\mathbf{o}_j$ and $\mathbf{r}_j$ the body's rest offset
and quaternion relative to its parent. Let $\bar{\mathbf{c}}_j$ and
$\bar{\mathbf{a}}_j$ be the rest anchor and axis expressed in the robot base frame. We form
\begin{equation}
\begin{aligned}
\mathbf{v}_j=\big[&\mathbf{a}_j,\;\mathbf{c}_j/s_R,\;\mathbf{o}_j/s_R,\;\mathbf{r}_j,
                  \;\|\mathbf{o}_j\|_2/s_R,\\
                 &\ell_j,\;u_j,\;u_j-\ell_j,\;d_j/10,\;\eta_j,\;\log s_R,\\
                 &\bar{\mathbf{c}}_j/s_R,\;\bar{\mathbf{a}}_j,\;m_j\big],
\end{aligned}
\label{eq:descriptor}
\end{equation}
where $[\ell_j,u_j]$ is the joint range, $d_j$ the depth in the compiled body tree,
$s_R$ the robot's zero-pose vertical extent measured from body origins in meters, and
$\eta_j=\operatorname{sign}(\bar c_{j,y})$ a base-frame side indicator, with a small
zero tolerance. The flag $m_j$ indicates whether the joint has finite limits.
A shared multilayer perceptron (MLP) maps the $27$-dimensional descriptor to a query $\mathbf{t}_j=\phi(\mathbf{v}_j)$.
Local geometry describes how a joint moves, while the base-frame anchor, axis and
side fields encode its spatial identity, distinguishing left--right counterparts
and otherwise symmetric joints without relying on joint names.
Normalizing offsets expresses geometry relative to body size, while $\log s_R$
preserves absolute scale. For a new embodiment, we supply its joint descriptors
and output ranges while reusing all learned weights.

\myparagraph{Visual features and shared decoder.} PromptHMR's frozen image and
camera encoders~\cite{promlthmr} provide visual features conditioned on crop intrinsics.
We pool features to $32\times32$ image tokens $\mathbf{X}$ and initialize the decoder as
\begin{equation}
\mathbf{H}^{0}=
[\mathbf{e}_{\rm root},\mathbf{e}_{\rm aux},\mathbf{t}_1,\ldots,\mathbf{t}_n],
\end{equation}
where the learned root and auxiliary tokens are shared across robots. The root
token predicts root pose. The auxiliary token has no separate prediction target.
Four pre-normalized transformer blocks combine token self-attention, cross-attention
to $\mathbf{X}$ and feed-forward updates with residual connections. A shared linear
head maps each normalized joint token to a scalar $z_j$, decoded as
\begin{equation}
\hat q_j=\ell_j+(u_j-\ell_j)\,\sigma(z_j).
\label{eq:bounded}
\end{equation}
Here $\sigma$ is sigmoid. For continuous joints, $\hat q_j=\operatorname{wrap}(z_j)$.

\myparagraph{Camera-relative root pose.}
\label{sec:root}
Nominal robot height sets metric depth, and the crop-center ray sets a lateral
reference. A linear root-token readout predicts $6$D orientation~\cite{rot6d} and
dimensionless corrections $(p_x,p_y,p_z)$ to account for posture and image evidence.
Let $H_R$ denote robot height, $S$ the square crop size, $\rho$ the reference
robot-height fraction, and $(f_x,f_y),(c_x,c_y)$ the focal lengths and principal
point in crop pixels. With $u_0=S/2$, translation is
\begin{equation}
\begin{aligned}
\hat z&=\frac{f_xH_R}{S\rho}\exp\big(\operatorname{clip}(p_z,-3,3)\big),\\
\hat t_x&=\left(\frac{u_0-c_x}{f_x}+p_x\right)\hat z,\\
\hat t_y&=\left(\frac{u_0-c_y}{f_y}+p_y\right)\hat z.
\end{aligned}
\label{eq:ray}
\end{equation}
The offsets $p_x,p_y$ adjust the crop-center ray in normalized camera coordinates,
while $p_z$ corrects log-depth, preserving positive depth. The metric height prior
$H_R$ is distinct from the body-origin extent $s_R$ used to normalize joint descriptors.

\myparagraph{Training objective.}
\label{sec:impl}
We supervise joint coordinates, their induced body poses, and the root transform.
Let $\mathcal{B}_R$ be the robot's body set and
$P_R(\qvec)_b$ the position of body $b$ in the robot base frame. The objective is
\begin{equation}
\begin{aligned}
\mathcal{L}_q&=\frac{1}{n}\sum_j
  \left|\operatorname{wrap}(\hat q_j-q_j)\right|,\\
\mathcal{L}_{\rm FK}&=\frac{1}{|\mathcal{B}_R|}\sum_b
  \left\|P_R(\hat{\qvec})_b-P_R(\qvec)_b\right\|_2,\\
\mathcal{L}&=\mathcal{L}_q+\lambda_{\rm FK}\mathcal{L}_{\rm FK}
 +\lambda_{\rm end}\mathcal{L}_{\rm end}\\
 &\quad+\lambda_{\rm root}\left(\mathcal{L}_{\rm rot}
       +\frac{\|\hat{\mathbf{t}}-\mathbf{t}\|_1}{\tau}\right).
\end{aligned}
\end{equation}
The angle difference is wrapped to $(-\pi,\pi]$, $\mathcal{L}_{\rm rot}$ is geodesic
rotation error, and $\tau=5$\,m scales translation loss. Differentiable
FK~\cite{horopose,dkpr,robotag} propagates body-position errors to joint predictions
in the robot base frame, separating articulation from root motion.
For terminal actuated bodies, $\mathcal{L}_{\rm end}$ averages
$1-\langle\hat{\mathbf{u}}_b,\mathbf{u}_b\rangle^2$, where the unit quaternions
$\hat{\mathbf{u}}_b,\mathbf{u}_b$ are predicted and target FK orientations. This
supervises rotations that leave terminal body origins unchanged.
We set $\lambda_{\rm FK}=1$, $\lambda_{\rm end}=0.1$ and $\lambda_{\rm root}=0.5$.

\myparagraph{Implementation.} The decoder has width $256$, eight attention heads and approximately $4.55$M active
parameters. We train with AdamW, learning rate $4\times10^{-4}$, effective batch size $48$ and a
one-cycle schedule, keeping the visual encoder frozen and grouping each device's
mini-batch by embodiment. Training uses $8000$ synthetic images per
embodiment for twenty epochs, selecting the checkpoint by the mean synthetic-validation
objective. The root decoder uses reference crop size $S=480$, with intrinsics rescaled
from encoder pixels to this crop resolution.
Nominal robot heights are estimated from zero-pose geometry bounds using a common
scale factor calibrated to G1's known height.

\subsection{World-Space Motion Recovery}
\label{sec:world}
To express the recovered poses in a common world frame, we combine each predicted
root-to-camera transform with the camera-to-world transform. Here
$\mathbf{T}_{A\leftarrow B}$ maps coordinates from frame $B$ to frame $A$:
\begin{equation}
\hat{\mathbf{T}}_{W\leftarrow R,t}
=\mathbf{T}_{W\leftarrow C,t}\,\hat{\mathbf{T}}_{C\leftarrow R,t}.
\end{equation}
Together with $\hat{\qvec}_t$, this transform places the articulated robot in world
coordinates through forward kinematics. For a fixed camera, the camera-to-world
transform is constant. For a moving camera, we mask the robot to exclude its motion
from camera estimation and use the PromptHMR camera pipeline~\cite{promlthmr} with
DROID-SLAM~\cite{droidslam}. Metric depth supplies the translation scale. The world
frame is defined independently for each continuous shot.

\section{Experiments}
Our experiments assess \humanoidPose{} as an external source of robot-state
information, evaluating its ability to recover joint configurations and global
motion from visual observations. We examine accuracy and generalization across
embodiments and domains, the contribution of the main design choices, and
adaptation with additional visual data.

\begin{table*}[t]
\centering
\caption{\textsc{Pose estimation on synthetic data and real robots.}
MPJAE ($^\circ$) and base-frame MPJPE (cm).
All~(16) averages robots equally. One \humanoidPose{} model serves all robots and both domains.
Best in bold and second best underlined.
}
\label{tab:pose_results}
 
\setlength{\tabcolsep}{3pt}
\renewcommand{\arraystretch}{1.1}
\resizebox{0.95\textwidth}{!}{%
\begin{tabular}{lcccccccccccccc}
\toprule
\multirow{3}{*}{\textbf{Method}}
 & \multicolumn{8}{c}{\textbf{Synthetic}}
 & \multicolumn{6}{c}{\textbf{Real Robot}} \\
\cmidrule(lr){2-9}
\cmidrule(lr){10-15}
 & \multicolumn{4}{c}{$\mathrm{MPJAE}\,\downarrow_{^\circ}$}
 & \multicolumn{4}{c}{$\mathrm{MPJPE}\,\downarrow_{\mathrm{cm}}$}
 & \multicolumn{3}{c}{$\mathrm{MPJAE}\,\downarrow_{^\circ}$}
 & \multicolumn{3}{c}{$\mathrm{MPJPE}\,\downarrow_{\mathrm{cm}}$} \\
\cmidrule(lr){2-5}
\cmidrule(lr){6-9}
\cmidrule(lr){10-12}
\cmidrule(lr){13-15}
 & All (16) & G1 & PM01 & N1
 & All (16) & G1 & PM01 & N1
 & G1 & PM01 & N1
 & G1 & PM01 & N1 \\
\midrule
 
Keypoint fitting
 & 21.21 & 19.86 & 27.86 & 20.32
 & 10.84 & 9.85 & 15.27 & 8.74
 & 10.52 & \underline{6.99} & 14.70
 & 5.93 & 6.14 & \underline{5.63} \\
PromptHMR~\cite{promlthmr} + GMR~\cite{gmr}
 & 18.03 & 20.37 & 24.05 & 18.26
 & 8.78 & 10.17 & 12.82 & 7.45
 & 12.50 & 16.11 & 19.57
 & 5.31 & \underline{6.02} & 7.46 \\
 
PromptHMR~\cite{promlthmr} + GMR~\cite{gmr} + SynFT
 & 16.82 & 19.39 & 22.83 & 17.04
 & \underline{7.98} & 10.03 & 12.85 & \underline{6.99}
 & \underline{10.46} & 16.45 & 15.42
 & \underline{4.88} & 11.88 & 6.33 \\
HoRoPose~\cite{horopose}
 & 15.09 & 14.21 & \underline{20.60} & 16.19
 & 9.61 & 10.05 & \underline{12.06} & 8.91
 & 12.98 & 15.36 & 13.72
 & 13.24 & 11.01 & 7.24 \\
RoboPEPP~\cite{robopepp}
 & \underline{14.49} & \underline{13.62} & 20.91 & \underline{15.86}
 & 8.68 & \underline{8.84} & 14.23 & 8.06
 & 10.72 & 13.96 & \underline{10.52}
 & 7.97 & 8.40 & 6.00 \\
\midrule
 
\rowcolor{oursrow}
\textbf{\humanoidPose{} (Ours)}
 & \textbf{9.48} & \textbf{9.03} & \textbf{11.15} & \textbf{8.80}
 & \textbf{4.49} & \textbf{4.99} & \textbf{5.57} & \textbf{3.66}
 & \textbf{7.10} & \textbf{5.65} & \textbf{8.62}
 & \textbf{3.81} & \textbf{2.60} & \textbf{5.18} \\
\bottomrule
\end{tabular}%
}
\end{table*}

\begin{figure*}[t]
\centering
\includegraphics[width=\linewidth]{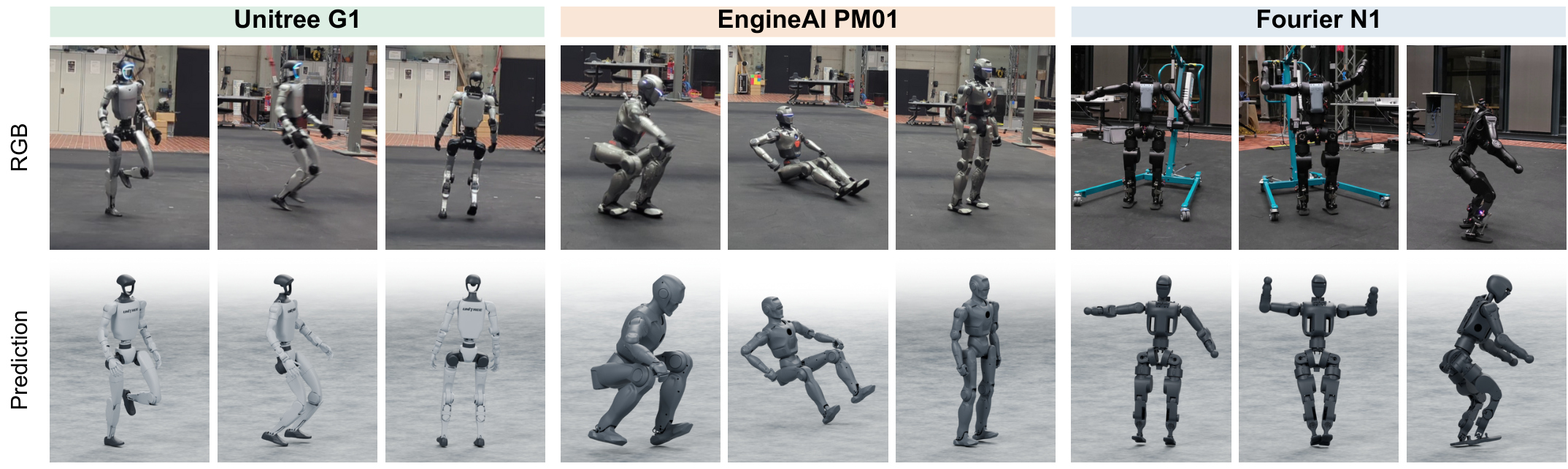}
\caption{Pose recovery across real humanoid platforms.
Self-recorded RGB frames (top) and predicted 3D configurations (bottom) for G1, PM01 and N1.
The same \humanoidPose{}  model recovers locomotion, arm movements, crouching and
supported poses across different robot structures and appearances.}
\vspace{-10pt}
\label{fig:qual}
\end{figure*}

\subsection{Experimental setup}
\myparagraph{Data and metrics.} We use our sixteen-robot synthetic dataset
(Sec.~\ref{sec:data}) with separate training and test images.
Real evaluation uses held-out, encoder-synchronized videos of Unitree G1, EngineAI PM01 and Fourier N1 performing locomotion, arm movements, crouching and supported poses, with synchronized optical motion capture on G1 for global root motion.
Mean per-joint angle error (MPJAE, $^\circ$) averages absolute wrapped joint-angle errors.
Mean per-joint position error (MPJPE, cm) averages Euclidean FK body-origin errors
in the robot base frame, with both root poses fixed to zero translation and identity rotation.
We pool real frames within each robot and average synthetic metrics equally across robots.
Global root translation is evaluated separately in Sec.~\ref{sec:rootreal}.

\myparagraph{Baselines.} Since full-body 3D humanoid pose estimation from RGB
images or videos remains underexplored, we construct and adapt the following
baselines for this task: PromptHMR~\cite{promlthmr} followed by GMR~\cite{gmr},
the same pipeline with synthetic finetuning (SynFT) on our data, keypoint fitting,
HoRoPose~\cite{horopose}, and RoboPEPP~\cite{robopepp}.

Finetuning updates PromptHMR's SMPL-X decoder
and camera embedding with its image encoder frozen. The training configuration
for \humanoidPose{} is given in Sec.~\ref{sec:impl}.
Synthetic finetuning uses the corresponding AMASS source labels for supervision.
Synthetic retargeting receives the source global pose and body shape. On real data, all
methods use square crops with robot-height fraction $0.55$, and GMR receives predicted
SMPL-X body rotations with a fixed canonical body shape and zero global orientation
and translation.
Retargeting solves each frame independently with the same GMR call for every method.

\emph{Keypoint fitting} is inspired by keypoint-based pose estimation~\cite{dream,robopepp}.
For each image, it fits joint angles and base pose
to the $15$ 2D keypoints predicted by our robot-adapted YOLO11 model. The optimization minimizes
confidence-weighted reprojection error through forward kinematics, subject to joint
limits, and starts from the zero configuration and a box-based translation estimate.
It shares detections, crops and frames with all other methods and requires no training.

We adapt HoRoPose~\cite{horopose} and RoboPEPP~\cite{robopepp}, originally evaluated
on robot manipulators, to humanoids and train a separate model for each robot using
their original code. HoRoPose retains its original losses and uses a joint-angle
branch registered for each humanoid. RoboPEPP pretrains its embedding on our
renders before finetuning.

\subsection{Comparisons with Baselines}
\label{sec:comparisons}
\myparagraph{Pose estimation on synthetic data.}\label{sec:synthetic} All methods
share a held-out synthetic test set excluded from training.
\humanoidPose{} achieves the lowest average MPJAE and MPJPE
(Table~\ref{tab:pose_results}), reducing MPJAE by
$35\%$ relative to RoboPEPP, the strongest joint-angle baseline on synthetic data.
The shared model also leads both metrics on each reported embodiment,
supporting recovery across robot structures.

\myparagraph{Real-robot pose estimation.}\label{sec:realbench}\label{sec:indomain}
The shared decoder transfers to all three real platforms without real pose
supervision, achieving the lowest MPJAE and MPJPE on each platform
(Table~\ref{tab:pose_results}). Relative to the best-performing baseline on each
platform, MPJAE decreases by $18$--$32\%$, with the largest gain on G1.
Figure~\ref{fig:qual} shows recovered locomotion, arm movements and supported
poses across robot geometries, including side and rear views.

\myparagraph{Root and world-space recovery.}\label{sec:multireal}\label{sec:rootreal}
Table~\ref{tab:root} evaluates global root position against motion capture on
G1 recordings captured with phone cameras. We compare against keypoint fitting
and PromptHMR with box prompts and camera intrinsics. PromptHMR is evaluated at
native scale and with translation scaled per sequence by robot height divided by
median predicted mesh height. Each trajectory receives one rigid
alignment without rescaling, refined on the lowest $90\%$ of initial residuals.
Errors are pooled over all matched frames shared by the methods. \humanoidPose{}
achieves the lowest error on all metrics and reduces root-mean-square error (RMSE) by $40\%$ relative to
height-scaled PromptHMR. Our root estimator uses known robot size and learns
image-dependent position corrections from synthetic robot poses, helping account
for posture changes beyond a fixed height-based scale adjustment.

\begin{table}[!ht]
\centering\footnotesize
\caption{Global root-translation errors on G1 (cm) after fixed-scale rigid alignment.
P90 denotes the $90$th percentile.}
\label{tab:root}
\setlength{\tabcolsep}{4pt}
\begin{tabularx}{\linewidth}{@{}l*{3}{>{\centering\arraybackslash}X}@{}}
\toprule
Method & RMSE$\,\downarrow$ & Median$\,\downarrow$ & P90$\,\downarrow$ \\
\midrule
Keypoint fitting & 42.2 & 21.3 & 56.3 \\
PromptHMR~\cite{promlthmr} & 66.2 & 37.4 & 96.2 \\
PromptHMR~\cite{promlthmr} + height scaling & 39.0 & 16.8 & 51.1 \\
\midrule
\rowcolor{oursrow}
\textbf{\humanoidPose{}~(Ours)} & \textbf{23.4} & \textbf{11.3} & \textbf{21.7} \\
\bottomrule
\end{tabularx}
\end{table}

\begin{figure}[t]
\centering
\includegraphics[width=\linewidth]{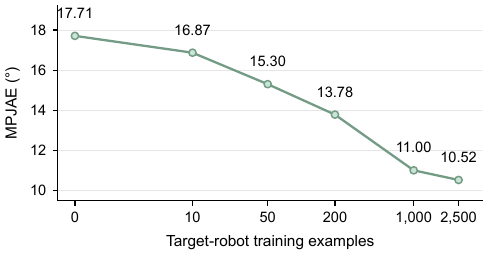}
\caption{Few-shot adaptation to a held-out embodiment.
G1 test error decreases with more synthetic training examples.
The horizontal axis uses $\log(1+N)$.}
\vspace{-10pt}
\label{fig:results}
\end{figure}

\subsection{Ablation studies}
\label{sec:ablations}
Table~\ref{tab:ablations} evaluates the main design choices on the same test
sets. Synthetic metrics average robots equally. Real metrics are frame-weighted
averages over G1, PM01 and N1.

\begin{table}[t]
\vspace{5pt}
\centering\footnotesize
\caption{Core design comparisons on synthetic data from sixteen robots and real data from three platforms.
MPJAE ($^\circ$) and MPJPE (cm).}
\label{tab:ablations}
\setlength{\tabcolsep}{2pt}
\begin{tabularx}{\linewidth}{@{}l*{4}{>{\centering\arraybackslash}X}@{}}
\toprule
 & \multicolumn{2}{c}{Synthetic} & \multicolumn{2}{c}{Real} \\
\cmidrule(lr){2-3}\cmidrule(lr){4-5}
Variant & MPJAE$\,\downarrow$ & MPJPE$\,\downarrow$ & MPJAE$\,\downarrow$ & MPJPE$\,\downarrow$ \\
\midrule
\rowcolor{oursrow}
\textbf{Ours} & \textbf{9.48} & \textbf{4.49} & \textbf{7.09} & \textbf{3.82} \\
\midrule
w/o joint identity & 10.94 & 5.05 & 7.15 & 4.02 \\
w/o direct robot decoding & 16.82 & 7.98 & 12.12 & 6.19 \\
w/o kinematic conditioning & 12.43 & 6.67 & 7.62 & 4.15 \\
w/o human-pretrained encoder & 11.14 & 5.57 & 7.95 & 4.65 \\
\bottomrule
\end{tabularx}
\end{table}

\myparagraph{w/o joint identity.} This variant removes the base-frame rest anchor
$\bar{\mathbf{c}}_j/s_R$, axis $\bar{\mathbf{a}}_j$ and side indicator $\eta_j$
from the joint descriptor in Eq.~\ref{eq:descriptor}. Errors increase in both
domains, with a larger effect on synthetic data. Base-frame geometry helps
associate visual evidence with the corresponding joints, including symmetric
left--right counterparts.

\myparagraph{w/o direct robot decoding.} This variant replaces direct prediction
of robot configurations with synthetic-finetuned PromptHMR followed by GMR
retargeting. This produces the largest increase in both errors among the compared
variants, highlighting the value of direct robot-state prediction.

\myparagraph{w/o kinematic conditioning.} This variant replaces the
descriptor-generated queries $\mathbf{t}_j=\phi(\mathbf{v}_j)$ with learned
embeddings for each robot's joints while retaining a shared
decoder. Both metrics worsen in both domains, supporting the combined use of
geometric descriptors.

\myparagraph{w/o human-pretrained encoder.} This variant replaces the
human-pretrained PromptHMR image encoder with a generic DINOv2~\cite{dinov2} encoder.
Both errors increase on synthetic and real data, indicating that human-pretrained
features provide a useful prior for robot articulation.

\subsection{Transfer to new embodiments}
\label{sec:cross}
We evaluate pose-decoder transfer while retaining the trained detector, whose
training includes the target robots' synthetic images and public videos.
We separately train decoders with a frozen PromptHMR encoder, holding out both G1
variants or PM01 and leaving fourteen or fifteen training robots. At test time,
the target's kinematic description supplies joint queries without learning new
embeddings. Pooled real MPJAE is $11.27^\circ$ on G1 and $11.44^\circ$ on PM01,
with median root errors of $8$--$16$\,cm across their sequences.

Finetuning the G1-held-out decoder on increasing amounts of synthetic G1 data
reduces synthetic test error (Fig.~\ref{fig:results}), with the image encoder
frozen and no real pose labels.

\subsection{Self-training from online videos}
\label{sec:selftrain}
Table~\ref{tab:detector} evaluates detector stages before and after self-training (ST).
On the sixteen-robot synthetic test set, percentage of correct keypoints (PCK)
uses a $5\%$ box-height threshold for visible landmarks.
Real coverage is the detection rate over all sampled frames with a visible robot.
Pose errors use frames detected by all three stages, with a fixed pose decoder.
The human-pose model's PCK uses the twelve shared landmarks. It detects robots on
G1 and PM01 but poorly localizes keypoints on N1. Synthetic finetuning improves
keypoint localization, and subsequent self-training improves real-video coverage
while retaining keypoint accuracy and yielding the lowest pose error across platforms.

\begin{table}[t]
\vspace{5pt}
\centering\footnotesize
\caption{Robot-adapted 2D pose detector stages: synthetic detection rate and PCK (\%), and real
coverage (\%) and MPJAE ($^\circ$) of our fixed pose decoder.}
\label{tab:detector}
\setlength{\tabcolsep}{1.5pt}
\resizebox{\linewidth}{!}{%
\begin{tabular}{@{}lcccccccc@{}}
\toprule
 & \multicolumn{2}{c}{Synthetic} & \multicolumn{2}{c}{G1} & \multicolumn{2}{c}{PM01} & \multicolumn{2}{c}{N1} \\
\cmidrule(lr){2-3}\cmidrule(lr){4-5}\cmidrule(lr){6-7}\cmidrule(lr){8-9}
Detector & Det.$\,\uparrow$ & PCK$\,\uparrow$ & Cov.$\,\uparrow$ & MPJAE$\,\downarrow$ & Cov.$\,\uparrow$ & MPJAE$\,\downarrow$ & Cov.$\,\uparrow$ & MPJAE$\,\downarrow$ \\
\midrule
YOLO11 & 56.8 & 38.6 & 98.6 & 7.82 & 96.5 & 5.89 & 77.8 & 17.65 \\
YOLO11 + SynFT & \textbf{100} & \textbf{93.6} & 81.1 & 8.65 & 75.6 & 5.87 & 99.3 & 8.98 \\
\rowcolor{oursrow}
YOLO11 + SynFT + ST & \textbf{100} & 92.1 & \textbf{98.8} & \textbf{7.11} & \textbf{96.8} & \textbf{5.36} & \textbf{99.4} & \textbf{8.90} \\
\bottomrule
\end{tabular}%
}
\end{table}

\section{Limitations}
\humanoidPose{} supports shared pose estimation across embodiments and transfer from
synthetic images to real observations. Our evaluation focuses on individual robots while
multi-robot scenes with sustained mutual occlusion remain to be explored. Such scenes
require maintaining robot identities and pose consistency as visual evidence becomes
intermittent, motivating temporal modeling and tracking. Downstream interaction and
learning from demonstrations require further evaluation to establish how pose
errors affect motion prediction, coordination and learned behaviors.

\section{Conclusion}
We presented \humanoidPose{}, a framework for recovering humanoid joint configurations
in world space from video. Combining pretrained human-pose features with
robot kinematics allows one model to learn across embodiments from synthetic
data, while self-training on unlabeled videos adapts the detector to real
appearances. Experiments
demonstrate improved pose accuracy over the evaluated baselines on real
humanoids and support transfer and adaptation to new embodiments. These findings
support vision as an external source of robot-state information when onboard
measurements are inaccessible. \humanoidPose{} provides a basis for understanding other humanoids
during interaction and for extracting internal states from online
and generated videos.
\section*{\small ACKNOWLEDGEMENTS}
{\small
The authors used generative AI to assist with manuscript editing and coding. All content was reviewed by the authors, who take full responsibility for it.
\par}

\bibliographystyle{IEEEtran}
\bibliography{refs}

\end{document}